\documentclass[letterpaper, 10 pt, conference]{ieeeconf}  

\usepackage{blindtext}
\usepackage{graphicx}
\usepackage{subfigure}
\usepackage{caption}
\usepackage{color}
\usepackage{subfig}
\usepackage{amsmath} 
\usepackage{amssymb}  
\usepackage{bm} 
\usepackage{tikz}
\usepackage{verbatim}
\usepackage{epstopdf}
\usepackage{tabularx}
\usepackage{url}
\usepackage{cite}
\usepackage{booktabs}
\usepackage{multirow}

\usepackage{algorithm}
\usepackage{algorithmicx}
\usepackage{algpseudocode}
\usepackage{amsmath}

\IEEEoverridecommandlockouts                              

\title{\LARGE \bf
Learning hierarchical behavior and motion planning for\\ autonomous driving
}

\author{Jingke Wang,
Yue Wang,
Dongkun Zhang,
Yezhou Yang, and
Rong Xiong
\thanks{Jingke Wang, Yue Wang, Dongkun Zhang and Rong Xiong are with the State Key Laboratory of Industrial Control and Technology, Zhejiang University, Hangzhou, P.R. China. Yezhou Yang is with School of Computing, Informatics, and Decision Systems Engineering, Arizona State University. Yue Wang is the corresponding author {\tt\small wangyue@iipc.zju.edu.cn}.}%
}

\begin{document}

\maketitle
\thispagestyle{empty}
\pagestyle{empty}

\begin{abstract}
Learning-based driving solution, a new branch for autonomous driving, is expected to simplify the modeling of driving by learning the underlying mechanisms from data. To improve the tactical decision-making for learning-based driving solution, we introduce hierarchical behavior and motion planning (HBMP) to explicitly model the behavior in learning-based solution. Due to the coupled action space of behavior and motion, it is challenging to solve HBMP problem using reinforcement learning (RL) for long-horizon driving tasks. We transform HBMP problem by integrating a classical sampling-based motion planner, of which the optimal cost is regarded as the rewards for high-level behavior learning. As a result, this formulation reduces action space and diversifies the rewards without losing the optimality of HBMP. In addition, we propose a sharable representation for input sensory data across simulation platforms and real-world environment, so that models trained in a fast event-based simulator, SUMO, can be used to initialize and accelerate the RL training in a dynamics based simulator, CARLA. Experimental results demonstrate the effectiveness of the method. Besides, the model is successfully transferred to the real-world, validating the generalization capability.
\end{abstract}

\section{Introduction}

Autonomous driving is an active research area in recent decades. The most widely adopted solution is the modular pipeline, which consists of route planning, behavior planning, and motion planning \cite{paden2016survey}. Route planning finds a global route in the road network to guide the vehicle to the goal based on the localization. Along the route, behavior planning makes tactical decisions, e.g. lane change, and motion planning accordingly yields motion trajectory. Both planners coordinately respond to the traffic situation, which is in general represented as a set of semantic objects via visual perception. The advantage of this solution is the interpretability. However, it requires highly accurate scene understanding and explicit modeling of all traffic scenarios, which impedes its wide application.

Learning-based driving solution is motivated by eliminating the requirements above. It leverages massive data to learn an implicit representation for a variety of traffic situations. A pioneering work of end-to-end imitation learning (IL) is proposed to directly map sensory data to final control command \cite{bojarski2016end}. The trained network acts as a combination of environment perception as well as behavior and motion planner, encoding the traffic situation as a hidden representation. These methods further inspire the incorporation of the global localization and routing so that the vehicle can reach arbitrary goals \cite{codevilla2018end}. To relieve the distribution mismatch caused by IL, reinforcement learning (RL) is employed to fine-tune the IL trained network \cite{liang2018cirl}, or learn driving from scratch \cite{perot2017end}. Results from these methods show good tracking of the global route, but weak tactical decision-making, causing the unawareness of traffic rules and limited maneuvers between road lanes.

\begin{figure}[tp]
\centering
\includegraphics[width=0.48\textwidth]{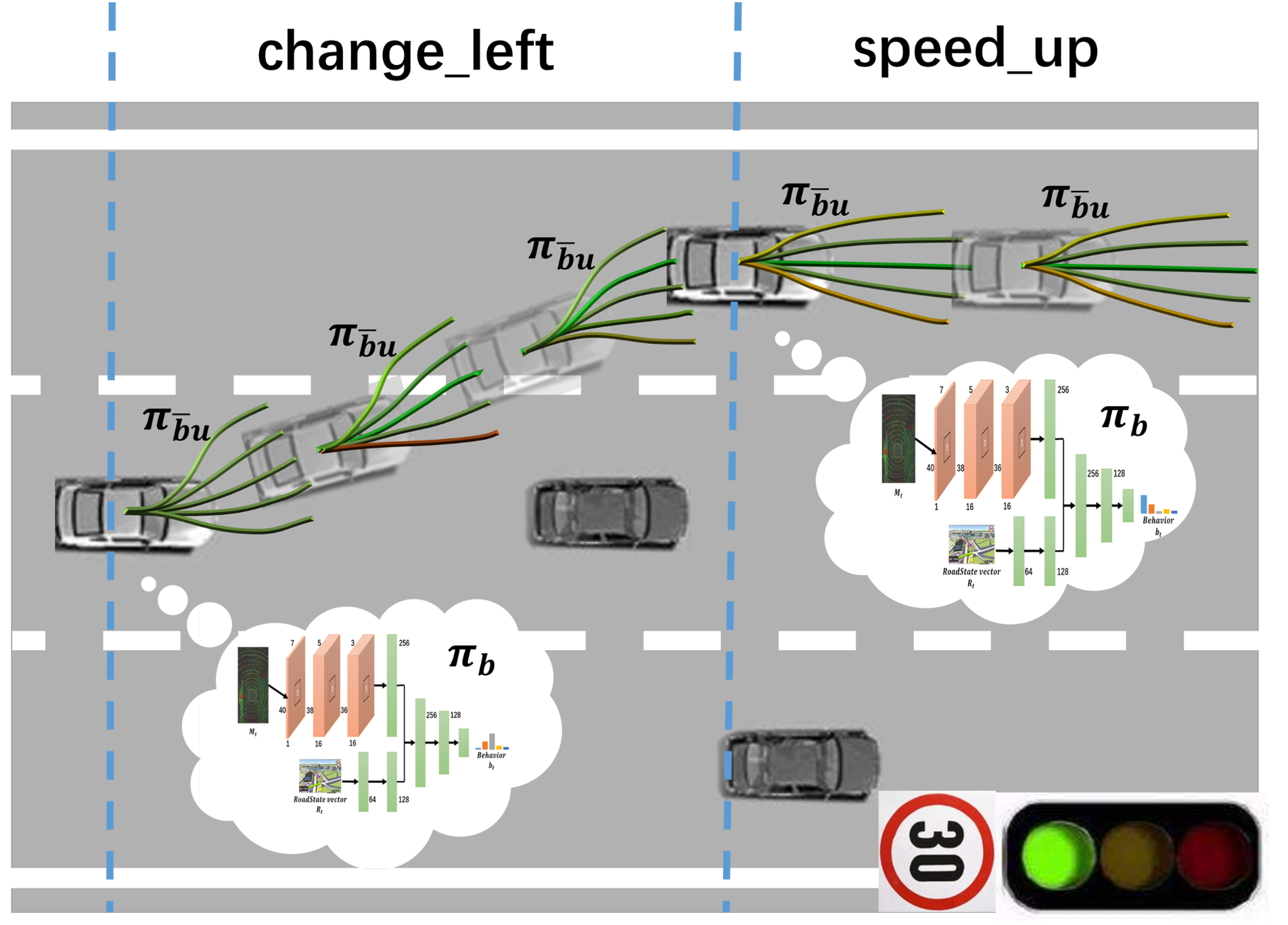}
\caption{HBMP for autonomous driving: $\pi_b$ is a high-level behavior planner yielding a tactical decision $\bar{b}$ for low-level motion planner $\pi_{\bar{b}u}$. The motion planner generates control command until finishing behavior $\bar{b}$. After that, a new $\bar{b}$ is yielded again.}
\label{teaser}
\vspace{-0.5cm}
\end{figure}

In end-to-end learning-based solution, a single network has to act as the whole modular pipeline. But only low-dimensional control commands or sparse rewards are utilized to supervise the learning. Due to this limitation, we consider that the intermediate behavior planner, which makes tactical decisions, might not be captured well in the learned model. Therefore, our idea is to introduce the hierarchical modeling of behavior and motion in the conventional modular pipeline to the learning-based solution.

Formally, we state the driving as hierarchical behavior and motion planning (HBMP) problem, which is shown in Fig. \ref{teaser}. The behavior layer makes a decision based on the current observation, e.g. lane change, while the motion planning layer yields the trajectory to complete this decision. Comparing with the existing works on learning behavior planning \cite{mukadam2017tactical,liu2018elements,shalev2016safe} using RL, we set to achieve a policy that optimizes both behavior and motion simultaneously. Considering the coupled action space of motion and behavior, the main challenge is the search inefficiency, especially in the long-horizon driving task. Therefore, we formulate HBMP problem as an equivalent high-level behavior learning problem. The key insight is assigning the optimal costs from low-level motion planner, i.e. a classical sampling-based planner, as the rewards for learning behavior. This formulation still gives the optimal solution to the HBMP problem but reduces the action space and diversifies the rewards, guaranteeing a better convergence. Besides, we initialize the RL policy network trained on CARLA \cite{dosovitskiy2017carla} using IL-based policy trained on SUMO \cite{behrisch2011sumo} by presenting a sharable sensory data representation, further accelerating RL training. Finally, the learned planner is transferred to new simulated and real-world environments to validate generalization capability. Our contributions are summarized as follow:
\begin{itemize}
\item We model the learning-based driving as HBMP to explicitly improve tactical driving. RL is applied to solve an equivalent behavior learning problem whose rewards are assigned by the cost from the motion planner. In this way, the final learned policy is still the optimal solution to HBMP.
\item We utilize an IL-based policy trained on SUMO to initialize the RL behavior policy network trained on CARLA based on sharable sensory data representation across simulation platforms and the real-world, accelerating the training and improving the transfer performance.
\item We validate the proposed planner on CARLA benchmark, showing superior performance in both training and testing scenarios. Besides, our method is able to actuate a real-world vehicle without fine-tuning the trained network.
\end{itemize}

\section{Related Works}

\subsection{Modular Pipelines}

Modular pipelines are the most mature autonomous vehicle solutions. They split the autonomous driving problem into perception, planning and control \cite{paden2016survey,schwarting2018planning}. Perception module parses the environment to sematic objects, which is fed to planning module to generate safe and efficient control command. A lot of teams adopt this framework in DARPA challenge \cite{thrun2006stanley, montemerlo2008junior}, and also for autonomous vehicle companies \cite{fan2018baidu}. Modular pipelines require very accurate environment parsing and various scenario modeling, calling for lots of manual labor. In addition, semantic objects based representation may be vulnerable to perception failure, and redundant considering the low dimensional control command, inspiring learning-based methods to explore potential intermediate representation \cite{chen2015deepdriving}.

\subsection{End-to-End Learning Methods}

An intuitive trial for learning-based driving is the end-to-end method. It directly maps sensory data to control command using a neural network. There are two lines of works: the imitation learning-based methods \cite{bojarski2016end, codevilla2018end, pan2017agile} and reinforcement learning-based methods\cite{sallab2017deep, wang2018deep, sallab2016end}.

The idea of imitation learning is to minimize the error between the model output and expert demonstration. In the early stage \cite{pomerleau1989alvinn}, these methods are tried on lane following and obstacle avoidance. More recently, global route and localization are also incorporated into the network so that the vehicle is able to reach arbitrary goals \cite{chen2019learning, codevilla2018end}. A problem in IL is the mismatch between training data and testing data. For example, in the experts' demonstration data, there is no data of driving towards the sidewalk. However, in the actual test, if the vehicle goes towards the sidewalk due to an error, the model takes unexpected actions because it has not seen such situations in the training data. Some efforts \cite{bojarski2016end, pan2017agile, kelly2019hg} are made in an attempt to address this problem, but it is still open \cite{codevilla2019exploring}.

Reinforcement learning is able to relieve the distribution mismatch. It follows a trial-and-error fashion. In recent years, we witness the success of RL on many tasks\cite{mnih2013playing, silver2017mastering}. Compared with these tasks, autonomous driving has a longer horizon and larger action space, making the efficiency of RL exploration a critical challenge. Some works integrate IL with supervision information from data level\cite{hosu2016playing, hester2018deep} or policy level \cite{liang2018cirl}, which provides guidance of exploration direction for RL. But learning complex behavior, e.g. maneuver, by RL remains a challenge.

The end-to-end methods reduce the requirements for perception and rule engineering by learning a representation of the traffic implicitly. However, to demonstrate a similar level of driving as the conventional modular pipeline, a single network has to model the perception, behavior and motion planner altogether. Given the low dimensional control commands as supervision in IL or handcrafted rewards in RL, it is likely that the network only learns the route tracking with limited tactical decision-making ability.

\subsection{Hierarchical Learning-Based Methods}

To reduce the task complexity for learning, some researchers propose to model the complex and long-horizon navigation task in hierarchy. There are various hierarchical structures. Affordance learning methods \cite{chen2015deepdriving, sauer2018conditional} use a network to estimate the human-designed affordance instead of semantic objects, while the planning is still based on traditional methods. These methods still employ an explicit handcrafted representation for traffic situations. There exist studies to learn a model from perception to behavior \cite{shalev2016safe, isele2018navigating, mukadam2017tactical} using RL based on a set of traffic rewards. These works separate different levels without joint optimization. In this paper, we follow the idea of HBMP in order to learn a representation of traffic situation implicitly and improve the tactical driving in learning-based solution. The difference is that we jointly optimize the motion and behavior in HBMP problem using RL.

\section{Problem Statement}
\label{ps}

Given the state $x_t$ consisting of ego vehicle and the traffic situation, and the control command $u_t$ at time $t$, we define the system model of the autonomous vehicle as
\begin{equation}\label{av}
  x_{t+1} = f(x_t,u_t).
\end{equation}
Note that we do not have the explicit form of (\ref{av}). The plan policy is defined as $u_t = \pi(x_t)$. The optimal planning of the control command with an assigned route is then stated as
\begin{equation}\label{opc}
  \mathop{\arg\max}\limits_{\pi} \gamma^T r_T + \sum_{t}\gamma^t r_t,
\end{equation}
where $r_T$ is the reward of the terminal state at end time $T$, and $r_t$ is the reward of the state trajectory during the time window $[0,T)$. The optimal plan policy maximizes the total reward in (\ref{opc}). In general, $r_T$ is specified based on whether the goal is reached, and $r_t$ penalizes the traffic rules breaking and energy or time consumption. As the problem is model-free, RL solver is utilized. Obviously, such reward design gives extremely sparse guidance only in the final stage, thus lots of trials may have almost the same rewards, causing the inefficiency of RL solver. Even in conventional methods based on explicitly semantic objects, the search is not very efficient.

To reduce the complexity, behavior planning is introduced to restrict the search space. Specifically, we denote a discrete variable $b_k$ as a high-level behavior. Then the plan policy is conditioned on $b_k$ as $u_t = \pi_{bu}(x_t,b_k)$. As the behavior is a tactical decision for a longer horizon than control command, each $b_k$ is fixed for a time interval $[t_k,t_{k+1})$. As shown in Fig. \ref{teaser}, $\{b_k\}$ partitions the timeline into multiple segments. In each segment, $b_k$ remains constant. Then, we have a partition of (\ref{opc}) as
\begin{equation}\label{obpc}
  \mathop{\arg\max}_{\pi_{bu}} \gamma^T r_T + \sum_k \gamma^{t_k} \sum_{t\in [t_k,t_{k+1})} \gamma^{t-t_k} r_t.
\end{equation}
We introduce a behavior policy as $b_k = \pi_b(x_{t_k})$. When a specific decision is made, say $\bar{b}$, $\pi_{bu}(x_t,b_k)$ becomes $\pi_{\bar{b}u}(x_t)$ during $[t_k,t_{k+1})$, transforming (\ref{obpc}) to
\begin{equation}\label{obpc2}
  \mathop{\arg\max}_{\pi_b}\{ \gamma^T r_T + \sum_k \gamma^{t_k} \max_{\pi_{\bar{b}u}}\sum_{t\in [t_k,t_{k+1})} \gamma^{t-t_k} r_t\},
\end{equation}
arriving at the HBMP formulation of the autonomous driving. Compared with the original formulation without explicitly modeling behavior (\ref{opc}), solving (\ref{obpc2}) can be divided into stages. The behavior decision only depends on the state. The conditional policy $\pi_{\bar{b}u}(x_t)$ has reduced search space when behavior is given. An intuitive example is that planning motion to a determined target lane, is simpler than planning both motion and target lane simultaneously. This factorization is popular in conventional methods, validating the feasibility \cite{paden2016survey}.

Suppose we have an optimal policy $\pi_{\bar{b}u}(x_t)$, then the reward is
\begin{equation}\label{obpc3}
  \mathop{\arg\max}_{\pi_b} \gamma^T r_T + \sum_k \gamma^{t_k} r_k,
\end{equation}
where $r_k=\max_{\pi_{\bar{b}u}} \sum_{t\in [t_k,t_{k+1})} \gamma^{t-t_k} r_t$, which is calculated by the optimal behavior conditioned control command $\pi_{\bar{b}u}(x_t)$.

We find that (\ref{obpc3}) has a similar form to (\ref{opc}) but has some important differences:

(i) In (\ref{opc}) the step of the timeline is uniform physical time while in (\ref{obpc3}), determined by the time for motion planner to finish $\bar{b}$. Therefore, (\ref{obpc3}) can be regarded as a reward function of a dynamic system with time delay on the behavior level as
\begin{equation}\label{avdelay}
  x_{t_{k+1}} = f_{\Delta t_k}(x_{t_k},b_k).
\end{equation}
Note that we have an approximation $\gamma^{t_k}=\gamma^k$ in (\ref{obpc3}) to arrive this result. As $\gamma$ is large, $\gamma=0.99$ in experiments, and the time interval for each behavior is short, the approximation does not break the convergence of RL.

(ii) The input $b_k$ in (\ref{avdelay}) is discrete, which has smaller space compared to the continuous $u_t$ in (\ref{opc}). With properly designed behavior, the RL search may begin with a feasible policy, though not optimal. It is critical as it yields higher chances for the positive reward.

(iii) In (\ref{obpc3}) the instant reward for the system (\ref{avdelay}) is the sum of optimal reward during multiple steps. It is likely to have more diverse rewards for different episodes, thus accelerating the convergence.

\section{Methodology and Implementation}
\label{mi}

Based on the discussion above, we propose two steps to solve HBMP problem using RL: (i) utilizing an optimal motion planner to achieve (\ref{obpc3}), (ii) adopting an initialization method to generate a feasible behavior planner. 

\subsection{State Modeling}

\begin{figure}[tp]
\centering
\includegraphics[width=0.41\textwidth]{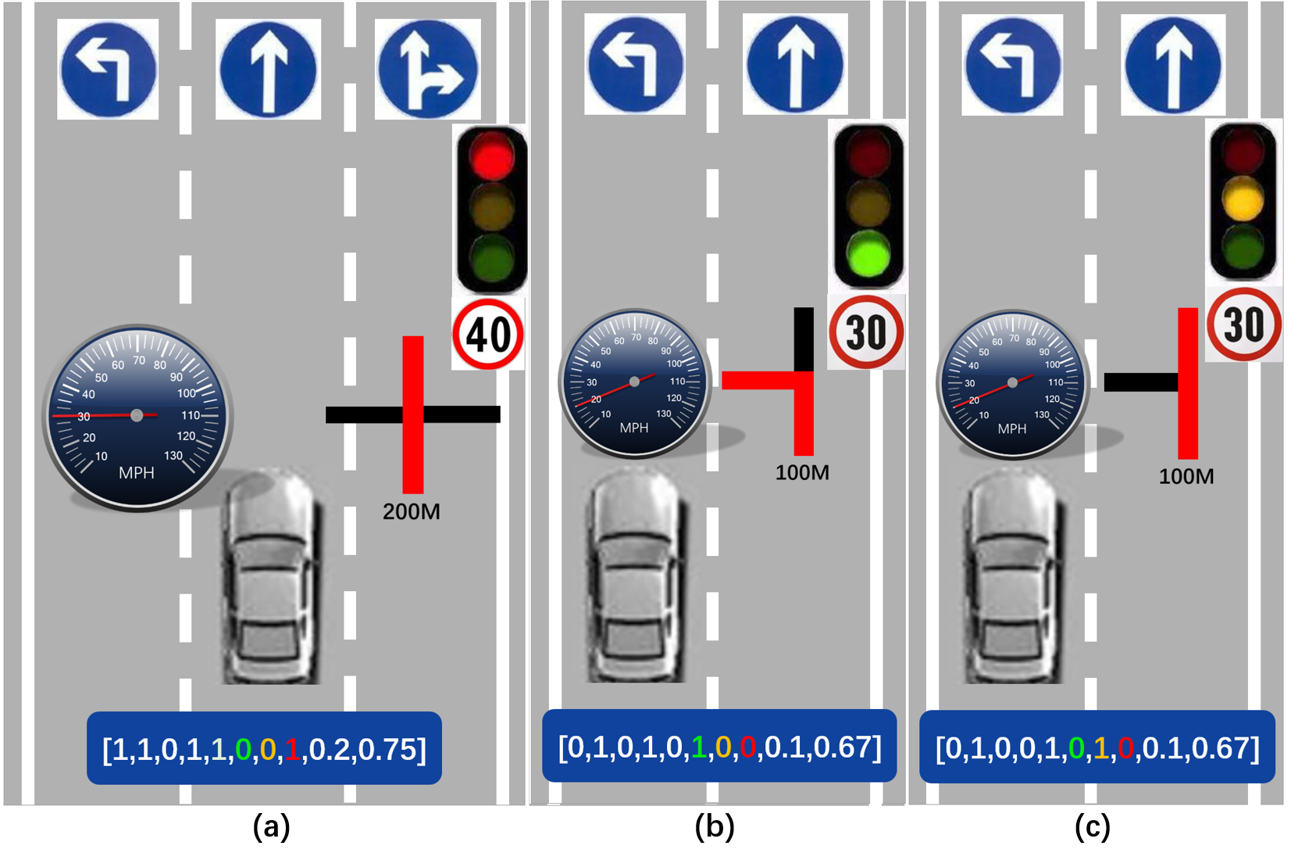}
\caption{Several cases of road state $R_t$. (a) Both left and right lanes exist, but only right two lanes can go straight based on route navigation. (b) Left lane does not exist and only the current lane can turn left. (c) Left lane does not exist and only the right lane can go straight. Suppose for all three graphs, the length of the road is $1000m$.}
\label{rp}
\vspace{-0.5cm}
\end{figure}

There are two parts of the information modeled in the system state. First, we use a vehicle-centric grid map $M_t$ to represent sensory data at each time step $t$. In each grid, the value is $1$ if there is an obstacle, otherwise it is $0$. Therefore, information from multiple range sensors can be integrated into a unified representation. This part of the information reflects the obstacles surrounding the vehicle. Note that no detection and recognition is utilized to parse the sensory data but only geometric transform calculation.

For single-lane task, $M_t$ and velocity $v_t$ are sufficient. While for the multi-lane environment with traffic rules, road profile is indispensable. Because the local sensory data cannot tell the traffic rules to constrain the planner, e.g. lanes direction. We crop a local vehicle-centric road map from the whole road map based on the localization. By parsing the local road map into profile, we have three types of information: (i) The existence of the left and right lane, and the direction of the left, current and right lane, denoted as $[e_l,e_r,\alpha_l,\alpha_c,\alpha_r]$. Here we include the current lane direction to tell whether the vehicle is in the correct lane to turn or go straight in the next intersection. Utilization of both existence and direction to encode a lane status is to distinguish whether a lane change is possible. (ii) The status of the next traffic lights and the distance to the stop line of intersection, denoted as $[l_g,l_y,l_r]$ (one-hot vector) and $d_s$. $d_s$ is normalized by the distance of the lane segment between two intersections. (iii) The current speed limit, which is normalized by the ratio of current velocity to speed limit, $\delta_v$. Concatenating the information together, we have the road profile $R_t$. Several cases and the assignment of $R_t$ are shown in Fig. \ref{rp}.

\subsection{Behavior Conditioned Motion Planner}

\begin{figure}[tp]
\centering
\includegraphics[width=0.25\textwidth]{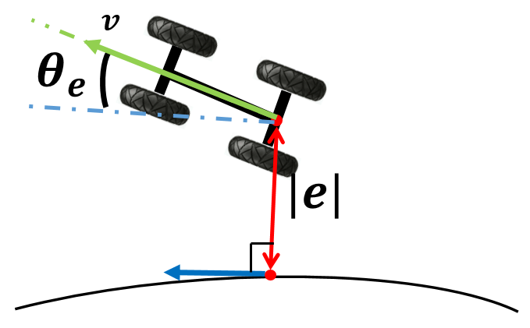}
\caption{The definition of lateral error $e$ and heading error $\theta_e$ on curvilinear coordinates. The blue arrow is the tangent vector of the curve.}
\label{cl}
\vspace{-0.5cm}
\end{figure}

We model 5 behaviors for the behavior planner as $b_k \in \{speed{\_}up,speed{\_}down,change{\_}left,change{\_}right,keep\}$. If the behavior conflicts with the $R_t$, $keep$ is made instead. When $speed\_up$ and $speed\_down$ are made, a reference linear velocity $v_{ref}$ is assigned by increasing or decreasing a percentage of the current feedback velocity. For $change\_left$, $change\_right$, and $keep$, $v_{ref}$ keeps the same to the current velocity. To achieve $v_{ref}$, we apply a linear acceleration velocity trajectory $v(t)$.

The vehicle has to follow the lane for all behaviors. We also build virtual lanes as in \cite{paden2016survey} so that $keep$ can achieve the turn by following the virtual lane. Based on the road map, we have the centerline of the target lanes as reference path, e.g. centerline of left lane when $b_k = change\_left$. As shown in Fig. \ref{cl}, we employ curvilinear coordinates to define the heading and lateral error $e$ and $\theta_e$. Using the controller in \cite{paden2016survey}, based on a timestep in $v(t)$, denoted as $v$, we have angular velocity $\omega$ as
\begin{equation}
\omega = \frac{v\kappa \cos\theta_e}{1-\kappa e}-(k_\theta|v|)\theta_e-(k_e v\frac{\sin\theta_e}{\theta_e})e,
\label{control}
\end{equation}
where $\kappa$ denotes the curvature of the reference lane and $k_\theta$, $k_e$ are control parameters. Following this process for several timesteps, we generate the angular velocity trajectory $\omega(t)$.

\begin{figure}[tp]
\centering
\includegraphics[width=0.5\textwidth]{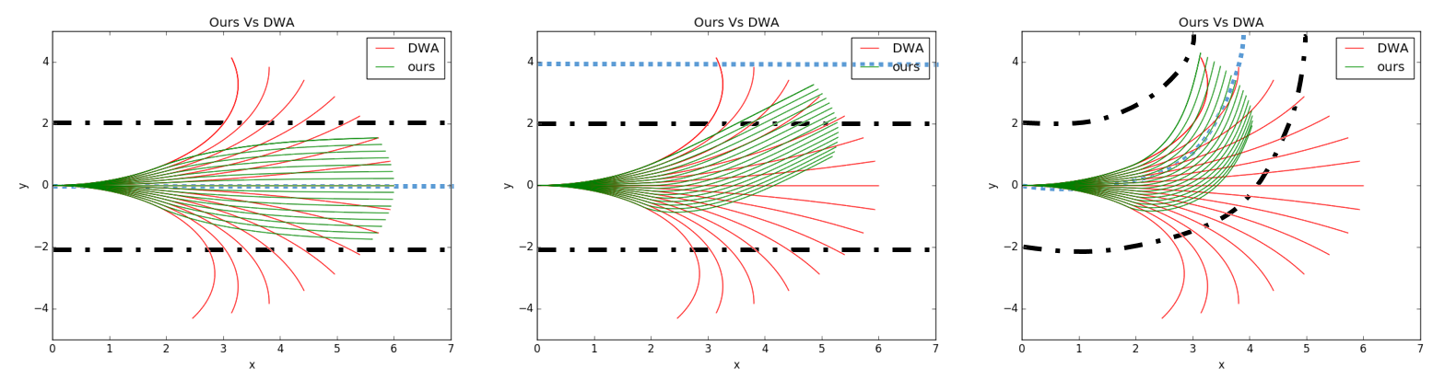}
\caption{Trajectories generated by our method (green) and DWA (red) during lane following, change left and turn left. The lane line is black and the centerline of target lane is blue.}
\label{mp}
\vspace{-0.5cm}
\end{figure}

To avoid obstacles, we follow the idea of dynamic window approach (DWA)\cite{fox1997dynamic} to span a set of initial linear and angular velocity samples by adding variations in $v(0)$ and $\omega(0)$. Then we generate a set of trajectories $\{v(t),\omega(t)\}$ following the methods above. Different from DWA, this method keeps all trajectories converging to the reference lane as shown in Fig. \ref{mp}. To select the best trajectory with the minimal cost, the cost function is designed as
\begin{eqnarray}
  c_{velocity} &=& \frac{\sum_{t}t^2|v_{ref}-v(t)|}{\sum_{t}t^2}\label{v},\\
  c_{dist} &=& \frac{1}{1+\sum_t |v(t)| }, \\
  c_{obs} &=& \sum_t d_{olon}(t) + \kappa d_{olat}(t),
\end{eqnarray}
where $d_{olon}$ and $d_{olat}$ are the longitudinal and lateral distance to the obstacles on the lanes and $\kappa$ is a weight. The total cost $c_{total}$ is the weighted sum of the three terms, encouraging a safe, low-energy and accurate trajectory. The selected trajectory is then send to the PID controller to actuate the steer, throttle and brake. When $e$ and $\theta_e$ is less than a threshold, the behavior finishes. Note that we plan the trajectory on $M_t$, which is an instant state of the environment without dynamics, so the trajectory may cause failures. We regard the failure as action failure and leave it to RL.

\subsection{Reward Definition}

As mentioned in Section \ref{ps}, reward engineering for RL is challenging, especially for long-horizon tasks. Based on the HBMP formulation, the high-level reward can be built by an aggregation of low-level rewards. Note that the underlying problem solved by the motion planner is to optimize $c_{total}$. We can reformulate the problem as
\begin{equation}\label{mpreward}
  \mathop{\arg\min}\limits_{\pi_{\bar{b}u}} c_{total} = \mathop{\arg\max}\limits_{\pi_{\bar{b}u}} \lambda( c_{total}),
\end{equation}
where $\lambda$ is a mapping with negative scaling. Recalling (\ref{obpc3}), we assign $\lambda( c_{total})$ of the executed trajectory in $[t_k,t_{k+1})$, denoted as $\lambda(\tilde{c}_{total})$, to the aggregated reward $\sum_{t\in [t_k,t_{k+1})} \gamma^{t-t_k} r_t$. We approximate $\gamma^{t-t_k}\approx 1$ since the behavior interval is short. As the trajectory is obtained by exactly maximizing the aggregated reward, we have
\begin{equation}\label{rkset}
  r_k = \max_{\pi_{\bar{b}u}} \lambda( c_{total}) =  \lambda( \tilde{c}_{total}).
\end{equation}
This reward design simplifies RL, since the sampling-based motion planner is optimal without training. Naturally, any optimal motion planner can be deployed as $\pi_{\bar{b}u}$. The only difference is that definition of $r_k$ should be accordingly determined by the cost function optimized in the chosen optimal motion planner.

For the final state reward $r_T$ in (\ref{opc}), we have
\begin{equation}\label{sparse}
r_T = \left\{\begin{array}{cc}
               100 & goal\_reached \\
               -50 & collision\_or\_overtime \\
               -10 & red\_light\_violation \\
               -1 & wrong\_lane
             \end{array}
\right..
\end{equation}
Using such rewards, the policy can be learned more efficiently as shown in the experiments in Section \ref{exp}.

\subsection{Initialization by IL}

To further improve RL efficiency, we train a behavior policy $\pi_b(x_{t_k})$ via IL for initialization. Instead of collecting demonstration data in the dynamics level simulator, CARLA, we demonstrate the behavior in SUMO, which is an event level simulator. In SUMO, the behavior is finished in one time step, e.g. the vehicle changes its lane one step after the decision is made. It is equivalent to a motion planner that can achieve the decision in minimal time. Hence demonstrating behavior in SUMO by human experts is very simple compared with CARLA. Denote the demonstration data as $\{x_k,\tilde{b}_k\}$, we use DAgger \cite{ross2011reduction} to train the IL policy loss
\begin{equation}\label{ilinit}
  \min_{\pi_{b}}\sum \|\tilde{b}_k - \pi_{b}(x_k)\|,
\end{equation}
where $\tilde{b}_k$ is the demonstrated behavior. We use the trained policy to initialize the network in RL.

\begin{figure}[tp]
\centering
\includegraphics[width=0.4\textwidth]{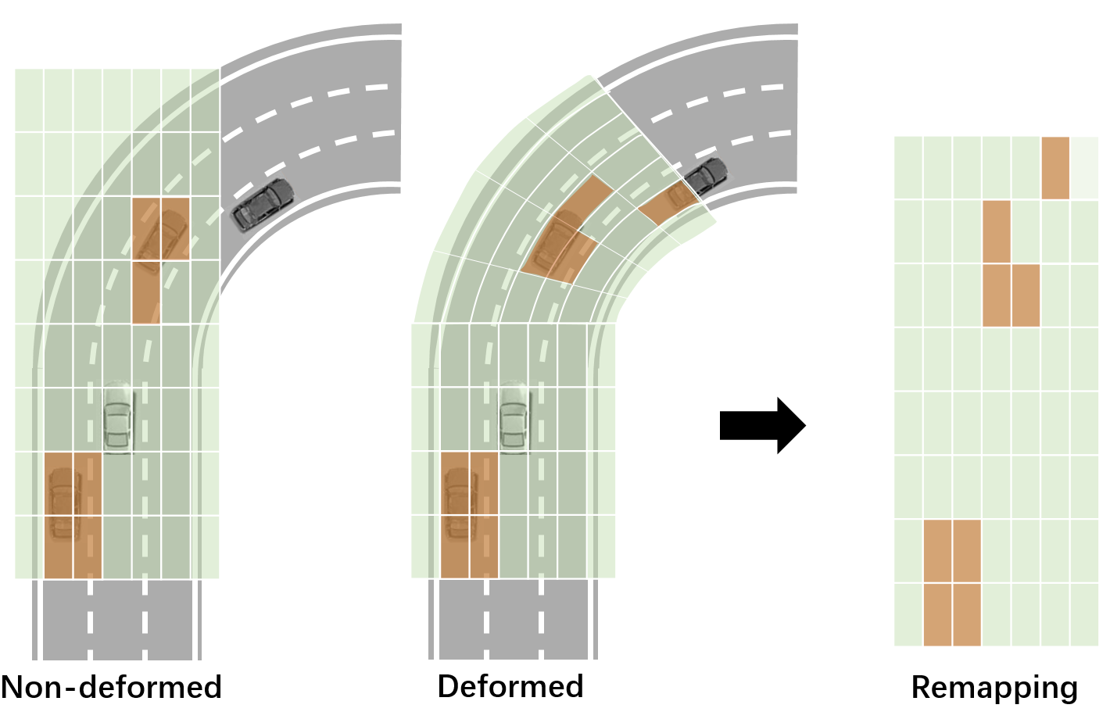}
\caption{An example of the sharable representation $M_t$. We build $M_t$ on a curvilinear coordinates defined by the lane which the vehicle is following. The grids occupied by the vehicle are marked $1$ (orange), otherwise $0$ (green).}
\label{deformM}
\vspace{-0.5cm}
\end{figure}

The road in SUMO is mostly straight and the heading of the vehicle is perfectly aligned with the lane direction. Therefore, $M_t$ built in SUMO is too ideal to generalize to roads in CARLA and the real-world. To tackle the problem, we improve $M_t$ using a curvilinear coordinate based definition as shown in Fig. \ref{deformM}. Note that the center of the deformed $M_t$ is on the lane instead of the vehicle, the deformed $M_t$ cannot reflect the lateral and heading error of the vehicle. Considering that the $M_t$ is utilized for behavior planning, the lateral and heading error at decision time are small, since a small error is the finish condition of the motion planner for the previous behavior. With the representation, we can achieve a good transfer to CARLA and real-world environment.

\subsection{Network and Training}

\begin{figure}[tp]
\centering
\includegraphics[width=0.5\textwidth]{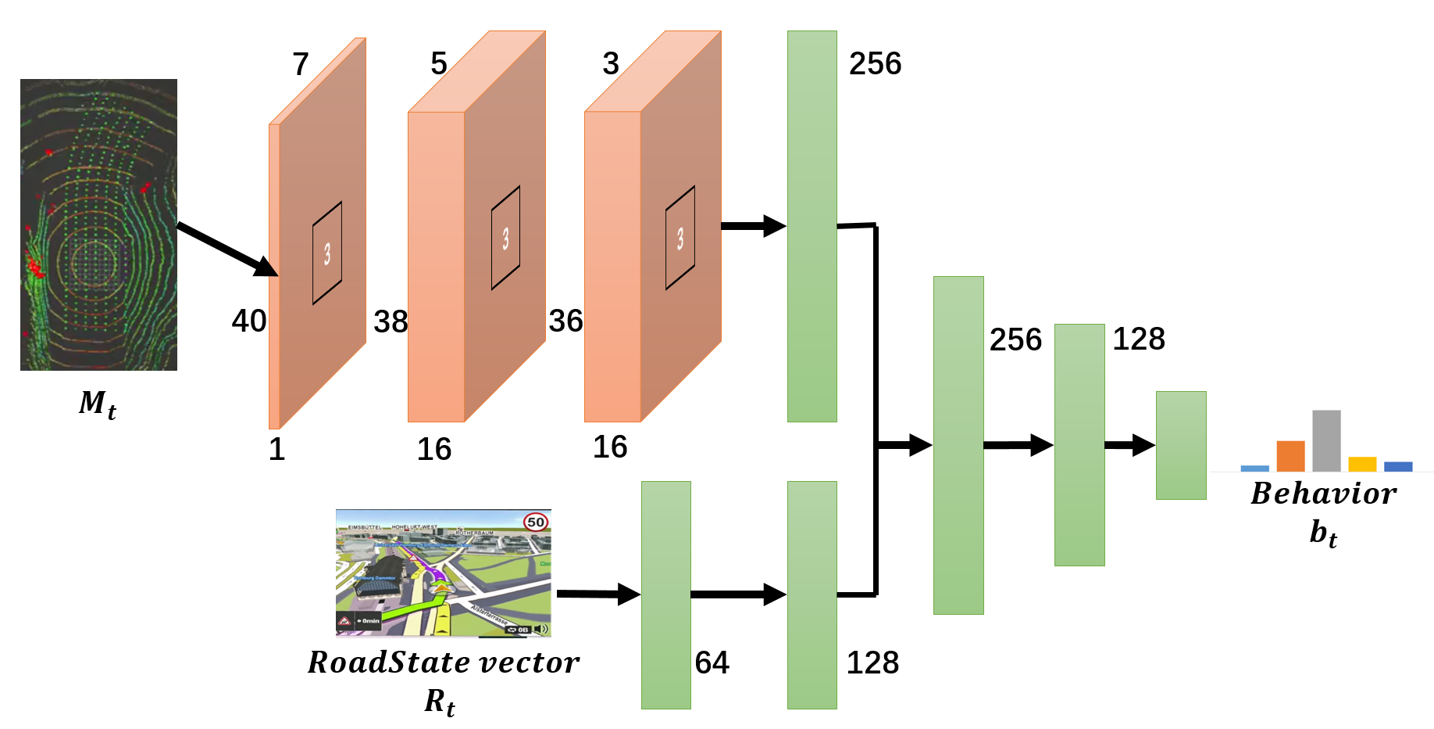}
\caption{The architecture of our actor-network with occupied map $M_t$, road state $R_t$ as input and behavior willing $b_t$ as output. $M_t$ and $R_t$ are fed into the two branches of the network and then concatenated and fed into FC. The architecture of the critic-network is the same as the actor-network except changing behavior output to state value.}
\label{network}
\vspace{-0.5cm}
\end{figure}

Finally, we report the implementation of the proposed method. We use actor-critic style Proximal Policy Optimization (PPO) \cite{schulman2017proximal} as the RL framework to learn a stochastic behavior policy $\pi_b(x_{t_k})$. The input of the network is the deformed map $M_t$ and the road profile $R_t$, the output of the actor-network is the distribution of the behaviors, while of the critic network is the state value. The architecture of the actor-network is shown in Fig. \ref{network}. The $M_t$ branch consists of three convolutional and one fully connected layers. The output of convolutional layers is flattened into the fully connected layers (FCs). The $R_t$ branch is multi-layer FCs. The outputs of two branches are concatenated and fed into the final FCs. Softmax is used after the output layer. The critic network has the same architecture except for the last layer, where the softmax is not assigned. The hyper-parameters of the network training are $Batchsize=32$, $LR_a=0.0001$, $LR_c=0.0005$ and $\epsilon_{CLIP}=0.2$.

\section{Experimental Results}
\label{exp}
We train our model on CARLA \cite{dosovitskiy2017carla}. Unlike most existing works training models in Town1, we train the model in Town3, which has richer road profiles e.g. multiple lanes. For each episode, the ego-vehicle is randomly born, and the goal is also randomly generated in the lane. $A^*$ is used to plan a global route as reference. Once the target is reached or the traffic rules are violated, an episode is terminated. We run $6000$ episodes to train the model.

The actor-network is trained on SUMO \cite{behrisch2011sumo} for initialization. To be similar to CARLA, we build a town with multi-lane roads with various traffic scenarios. In DAgger \cite{ross2011reduction}, the probability of choosing human output for control is $\alpha$ and the probability of choosing the model is $1-\alpha$. In the beginning, $\alpha=0.95$, and after $3000$ episodes of training, $\alpha$ gradually decreases to $0$.

The proposed method is implemented on Tensorflow \cite{abadi2016tensorflow}. In CARLA, the average frequency of decision-making is 1 Hz while the frequency of motion planning is 10 Hz. In SUMO, the frequency of decision-making is 10 Hz. The CPU for training is Intel i7-8700 and the GPU is GTX-1070.

\subsection{Ablation Studies}

\begin{figure}[tp]
\centering
\includegraphics[width=0.48\textwidth]{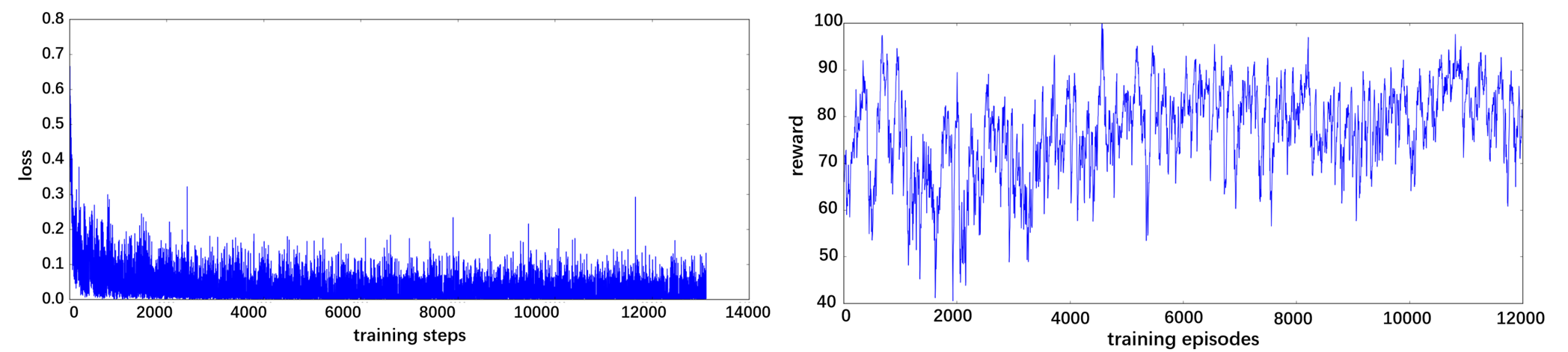}
\caption{The loss and reward curves for IL in SUMO using DAgger.}
\label{IL_loss_reward}
\vspace{-0.5cm}
\end{figure}

\begin{figure}[tp]
\centering
\includegraphics[width=0.48\textwidth]{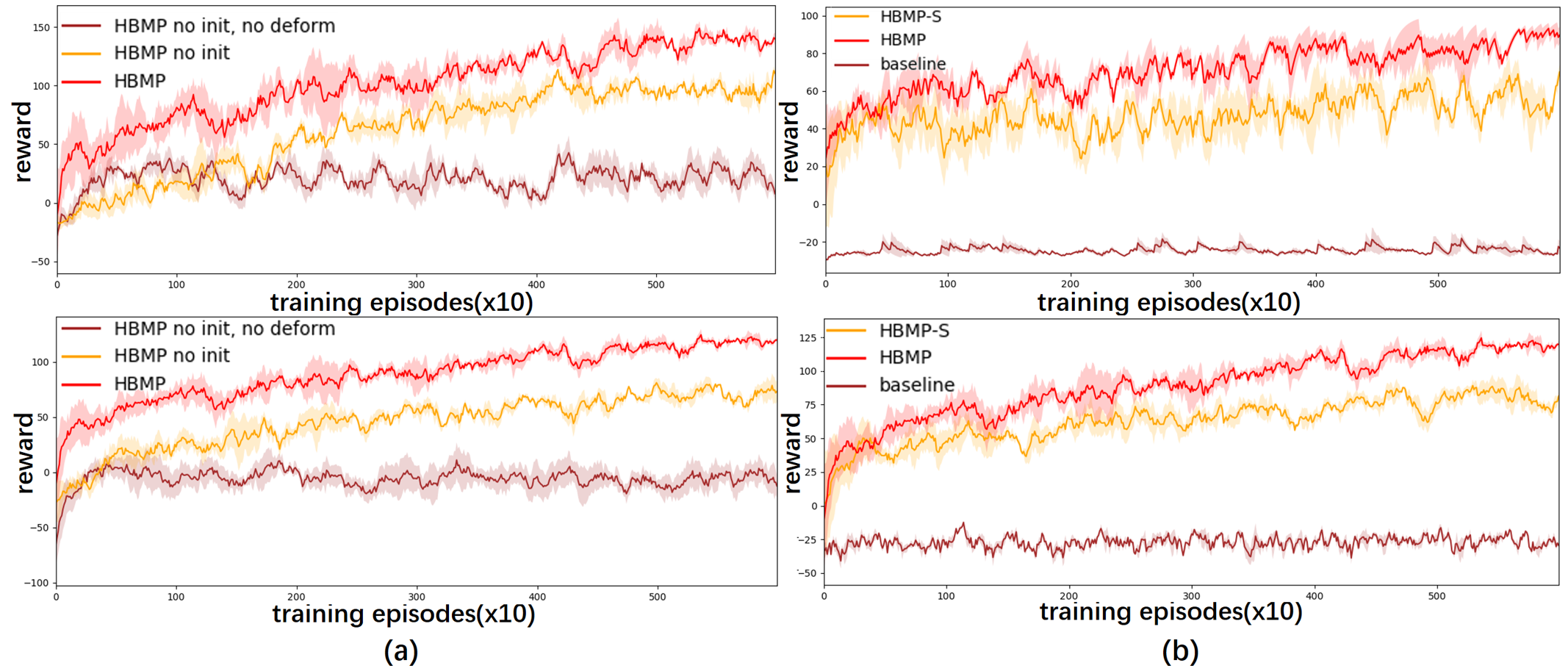}
\caption{The reward curve comparisons for different training methods evaluated by $r_T$ (upper row) and $r_T+\sum_t r_t$ (lower row): (a) RL policy with and without proper initialization, (b) RL policy with and without hierarchy as well as dense training reward.}
\label{ablation_result}
\end{figure}

\textbf{Initialization with IL.} Firstly, we show the loss and reward curves for DAgger in Fig. \ref{IL_loss_reward}. With the training, loss tends to converge. In the beginning, the reward is high, since most decisions are made by the human. The reward drops as the network decision is chosen with a higher probability. Then we see that the reward gradually turns back to a similar level to that at the beginning, demonstrating the convergence of the IL model. We use the trained model to initialize the actor-network in RL. In Fig. \ref{ablation_result}(a), we use $r_T$ and $r_T+\sum_t r_t$ as indicators to visualize the learning. With random initialization, the policy achieves increasing reward but constantly lower than that of SUMO initialized policy. SUMO initialized policies even keep increasing after $6000$ episodes. We find that the random initialized policy is almost a global route tracker without lane changes. In addition, with the proposed deformed $M_t$, the policy achieves higher reward than the non-deformed $M_t$, indicating the importance of deformed $M_t$ for transfer. Without deformation, the reward curves begin at a similar level to that of random initialized policy but increase later, showing that non-deformed representation may slightly contribute to the transfer.

\textbf{Effectiveness of HBMP.} We first validate the feasibility of $\sum_t r_t$ aggregated from motion planning. We still use the $r_T$ and $r_T+\sum_t r_t$ as indicators. As the two reward curves have similar trends in both Fig. \ref{ablation_result}(a) and (b), we know the two types of reward are correlated, suggesting that training with $r_T+\sum_t r_t$ is able to increase $r_T$.

Then we employ a baseline method, which directly maps input to $u$ without hierarchical modeling. It is trained using $r_T+\sum_t r_t$ except for (\ref{v}) as no reference velocity is made from behavior level. We see that almost no ascending trend is shown in Fig. \ref{ablation_result}(b). In comparison, HBMP policy with random initialization in Fig. \ref{ablation_result}(a) achieves significantly higher reward than the baseline method, showing the importance of hierarchical modeling to reduce the action space. By formulating the problem in hierarchy with a good initialization, we still have better convergence even only sparse reward $r_T$ is utilized for training (HBMP-S policy). Note that the proposed HBMP policy has the best performance, which owes to more diverse and denser rewards.

\begin{figure}[tp]
\centering
\includegraphics[width=0.48\textwidth]{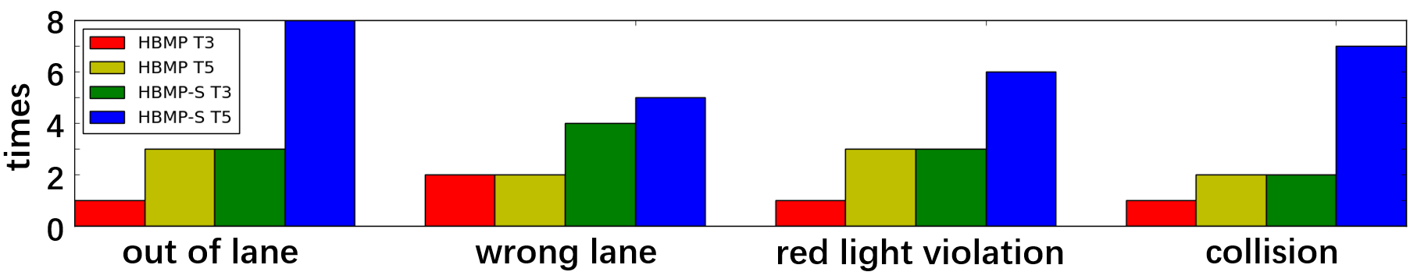}
\caption{Number of failures in $100$ times tests. In each group, four bars are for HBMP policy in Town3, in Town5, HBMP-S policy in Town3, in Town5, respectively.}
\label{traffic_rules}
\vspace{-0.5cm}
\end{figure}

\begin{figure}[tp]
\centering
\includegraphics[width=0.4\textwidth]{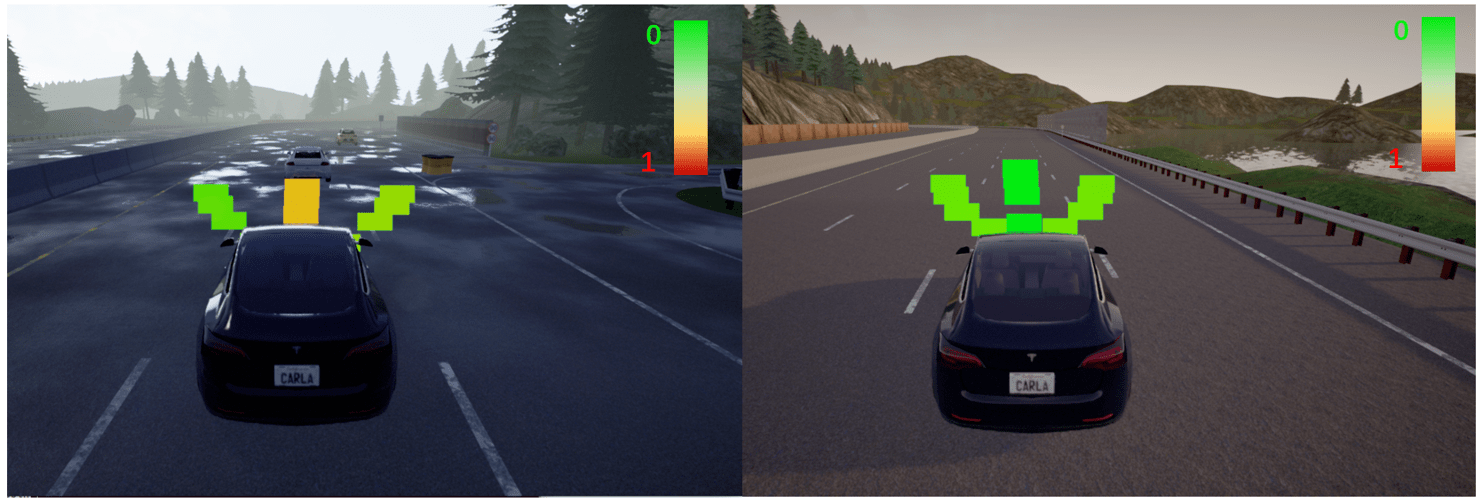}
\caption{The visualization of the cost from motion planner conditioned on $change\_left$, $keep$ and $change\_right$.}
\label{reward_visual}
\end{figure}

\textbf{Goal reach.} Furthermore, we evaluate HBMP and HBMP-S policies for goal reach in both training (Town3) and test map (Town5) as shown in Fig. \ref{traffic_rules}. We run each model for $100$ times. The results show that HBMP policy gives superior results in both training and test map compared with HBMP-S policy. To visualize the aggregated optimal reward from the motion planner, we show two cases in Fig. \ref{reward_visual}, from which we see that $r_k$ is reasonable to guide the RL.

\subsection{CARLA Benchmark Evaluation}
\begin{figure}[tp]
\centering
\includegraphics[width=0.49\textwidth]{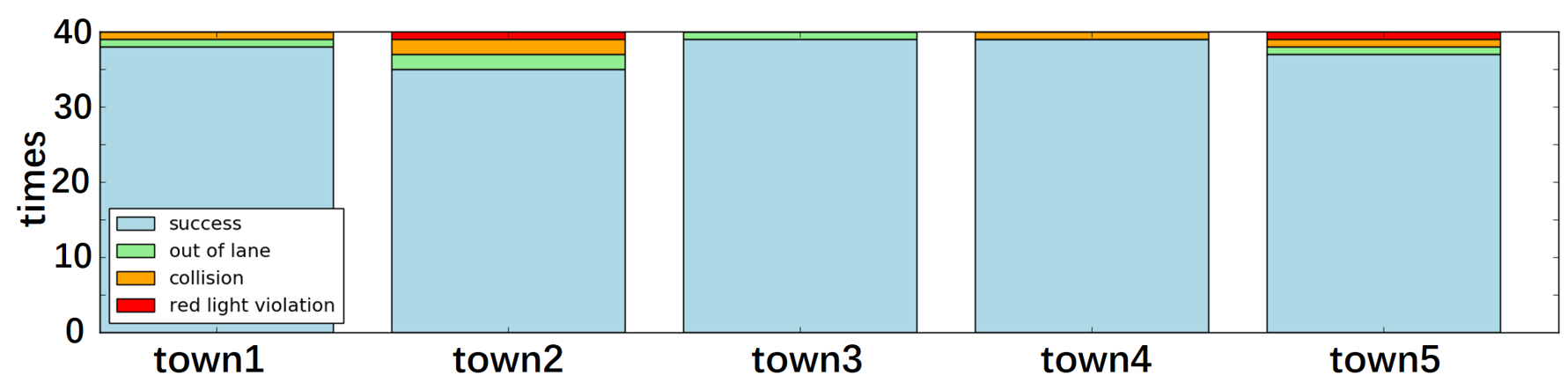}
\caption{Test results in $5$ towns. We test the model $40$ times in each town and count the times of successes, out of lanes, collisions, and red light violation.}
\label{town15}
\vspace{-0.5cm}
\end{figure}

\begin{figure}[tp]
\centering
\includegraphics[width=0.48\textwidth]{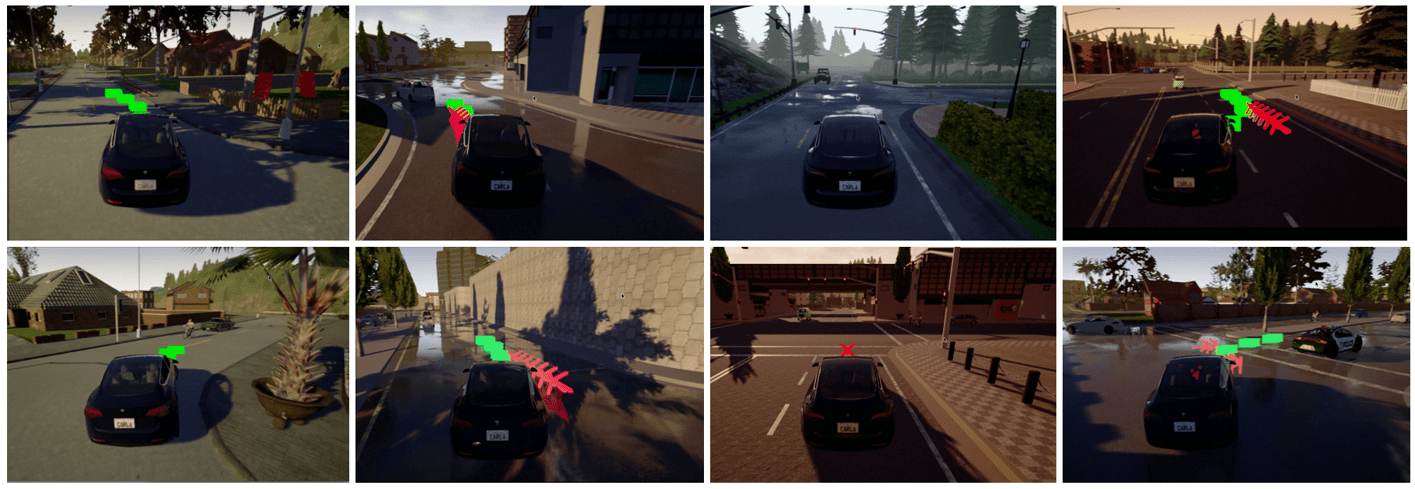}
\caption{Some driving cases in various CARLA environments including turning, single-lane driving, multi-lane driving, waiting for green light and overtaking, etc.}
\label{case_study}
\end{figure}

\begin{table*}[tbp]
\caption{Comparisons in Town1 \& Town2}
\begin{tabular}{c|p{0.45cm}<{\centering}p{0.45cm}<{\centering}p{0.45cm}<{\centering}p{0.45cm}<{\centering}p{0.45cm}<{\centering}p{0.45cm}<{\centering}p{0.45cm}<{\centering}p{0.45cm}<{\centering}p{0.45cm}<{\centering}|p{0.45cm}<{\centering}p{0.45cm}<{\centering}p{0.45cm}<{\centering}p{0.45cm}<{\centering}p{0.45cm}<{\centering}p{0.45cm}<{\centering}p{0.45cm}<{\centering}p{0.45cm}<{\centering}p{0.45cm}<{\centering}}
\hline
\multirow{2}{*}{\textbf{}} & \multicolumn{9}{c|}{Town1}                                  & \multicolumn{9}{c}{Town2 (New weather)}                                  \\ \cline{2-19}
                           & MP\cite{dosovitskiy2017carla} & IL\cite{dosovitskiy2017carla} & RL\cite{dosovitskiy2017carla} & CIL\cite{codevilla2018end} & CIRL\cite{liang2018cirl} & CAL\cite{sauer2018conditional} & MT\cite{li2018rethinking} & CILRS\cite{codevilla2019exploring} & ours         & MP\cite{dosovitskiy2017carla} & IL\cite{dosovitskiy2017carla} & RL\cite{dosovitskiy2017carla} & CIL\cite{codevilla2018end} & CIRL\cite{liang2018cirl} & CAL\cite{sauer2018conditional} & MT\cite{li2018rethinking} & CILRS\cite{codevilla2019exploring} & ours                 \\ \hline
Straight                   & 98 & 95 & 89 & 98  & 98   & 100 & 96 & 94    & \textbf{100} & 50 & 80 & 68 & 80  & 98   & 94  & 96 & 92    & \textbf{100} \\ \hline
One turn                   & 82 & 89 & 34 & 89  & 97   & 97  & 87 & 92    & \textbf{100} & 50 & 48 & 20 & 48  & 80   & 72  & 82 & 92    & \textbf{100} \\ \hline
Navigation                 & 80 & 86 & 14 & 86  & 93   & 92  & 81 & 88    & \textbf{100} & 47 & 44 & 6  & 44  & 68   & 68  & 78 & 88    & \textbf{100} \\ \hline
Nav. dynamic               & 77 & 83 & 7  & 83  & 82   & 83  & 81 & 85    & \textbf{95}  & 44 & 42 & 4  & 42  & 62   & 64  & 62 & 82    & \textbf{88}  \\ \hline
\end{tabular}
\label{comparisons_town12}
\end{table*}

\textbf{Comparative study.} We compare our HBMP policy with several state-of-the-art policies on CARLA benchmark. The model is tested 40 times on four tasks, i.e. Straight, One turn, Navigation and Navigation with dynamic obstacles in Town1 and Town2 respectively. The results are shown on Tab. \ref{comparisons_town12}. All methods except ours are trained in Town1 and tested in Town2, while our model is trained in Town3, so Town1 and Town2 are both test environments. In the results, our method achieves a $100\%$ success rate in static environment. Even dynamics present, our method still achieves a $95\%$ success rate in Town1 and $88\%$ success rate in Town2, much higher than other methods.

\textbf{Traffic indicators.} In order to further evaluate the generalization, we run 40 times random goal reach tasks for HBMP policy from Town1 to Town5 and count the number of goal reach, wrong lanes, collisions, and red light violation, as shown in Fig. \ref{town15}. Our model gets the highest success rate in Town3, which is the training environment. In general, our model achieves success rate more than $85\%$ in all maps, which verifies the generalization of our model. Some driving cases are shown in Fig. \ref{case_study}, including various driving tasks such as turning, single-lane driving, multi-lane driving, waiting for green light and overtaking.

\subsection{Real-World Vehicle Test}

\begin{figure}[tp]
\centering
\includegraphics[width=0.42\textwidth]{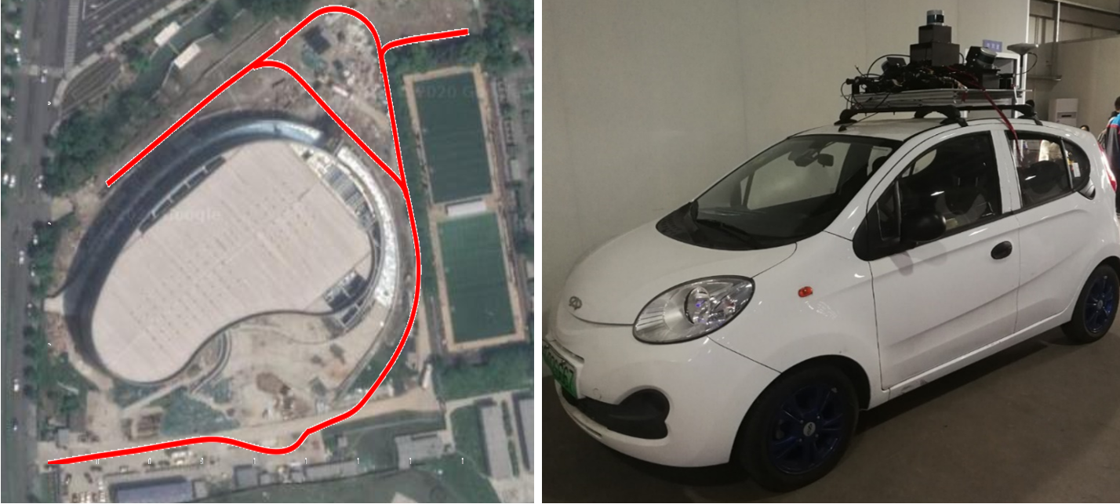}
\caption{Our real-world experiments setup. On the left is the road map in the campus satellite map. Right shows our test platform, a vehicle equipped with 3D LiDAR and RTK-INS.}
\label{realworld_setup}
\vspace{-0.5cm}
\end{figure}

The setup of our real-world experiments is shown in Fig. \ref{realworld_setup}. We built a road map of campus and plan a global route to guide the HBMP policy. The vehicle is equipped with RTK-INS for localization and 3D Lidar for perception. A laptop with Intel i5-9300H CPU and GTX-1650 GPU is used for onboard processing and sends the control signal to the vehicle CAN bus. The behavior network is directly transferred without tuning. We only tune some parameters of the motion planner to adapt to the vehicle.

The testing road has double lanes with a total length of $700m$,  including straight lanes, curve lanes, and intersections. We test our method in both natural and designed scenarios. There are few pedestrians and vehicles in the natural scenario. In the designed scenario, a cyclist is employed as a traffic participant to test the model in dynamic scenes. To ensure safety, we fix the maximal velocity as $15km/h$.

\begin{figure}[tp]
\centering
\includegraphics[width=0.49\textwidth]{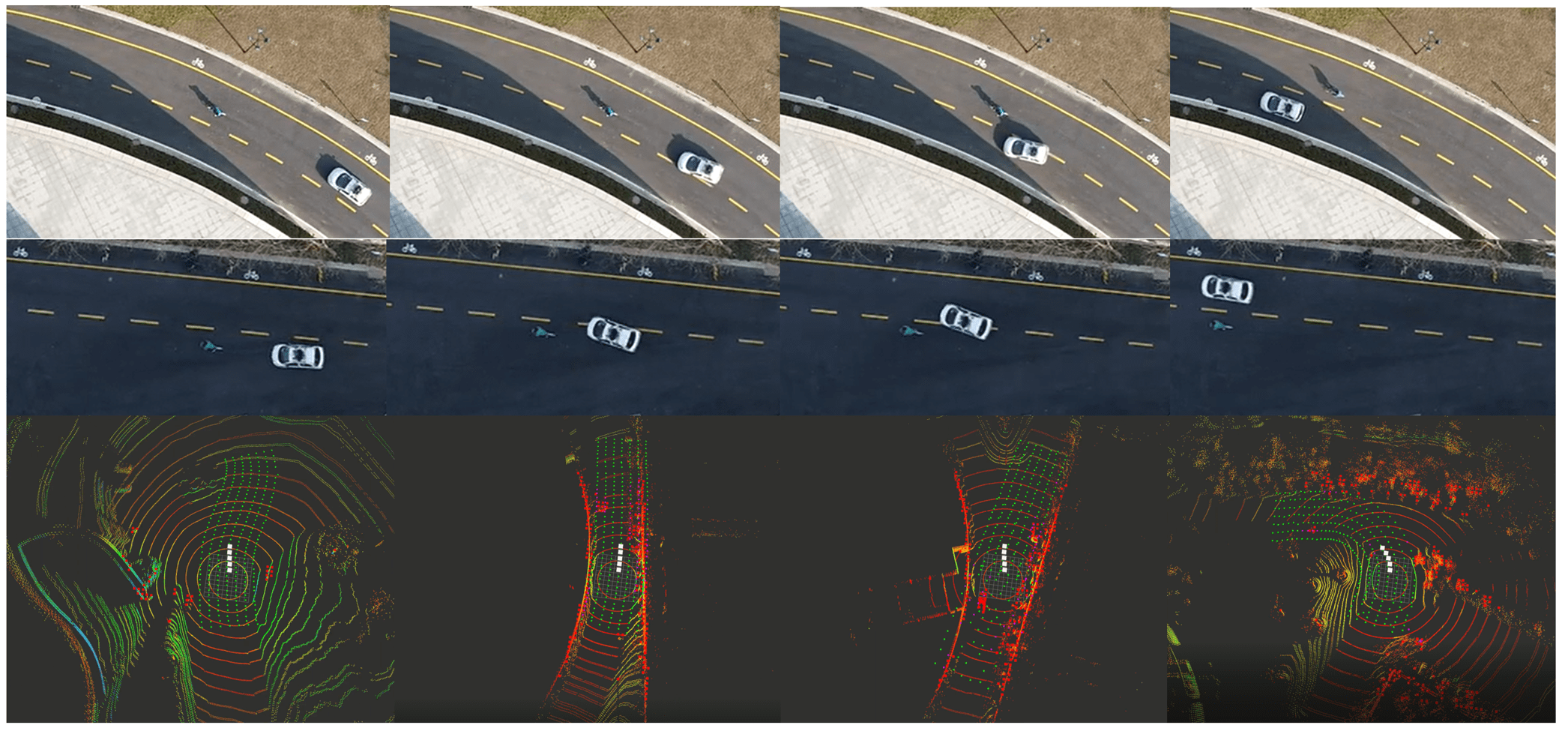}
\caption{The above two rows of pictures are the process of autonomous lane change overtaking from the perspective of aerial photography. The last row of pictures is the vehicle lidar vision and trajectory planning.}
\label{realworld_case}
\vspace{-0.5cm}
\end{figure}

No human intervention is delivered during the whole test. In the natural scene, since the ego-vehicle is slow, no overtaking is executed. The HBMP policy keeps the vehicle in the lane to track the global route in all shapes of lanes. In the designed scene, the vehicle overtakes the cyclist several times, validating the correct lane change decisions of the HBMP policy in the real-world. Two cases for $change\_left$ and $change\_right$ are shown in Fig. \ref{realworld_case}. Note that no cyclist recognition is utilized in the whole pipeline, but only occupied map directly extracted from lidar data. As the cyclist intentionally blocks the vehicle, $speed\_down$ and $speed\_up$ are made for many times since no such traffic participant exists in the simulation. Still, no abrupt braking is made.

\subsection{Discussion}

From the experiment, we find that HBMP policy performs well in simulation and generalization in real-world environment, which owes to the denser rewards from a low-level motion planner and good initialized policy based on deformed $M_t$. From the perspective of modular pipeline, global route planning and localization are regarded as input in many existing learning-based driving solutions, since these modules are relatively mature. We consider that learning motion planning may also be unnecessary as its maturity. Hence the proposed HBMP formulation provides a theoretic framework to couple results from classical motion planner into learning, resulting in joint optimality. For the generalization, sensory input representation $M_t$ is very important, which is also a usual representation in classical methods, naturally brings robustness across environments, enforcing the network to learn the decision-making, rather than extracting invariant scene features. As a summary, it can be a good choice to employ mature classical methods in the learning-based solution, leaving the network to focus on the problem that classical methods are not good at.

\section{Conclusion}

In this paper, we formulate the autonomous driving as a hierarchical behavior and motion planning problem. Then, we show that this problem is equivalent to an RL problem when the reward is an aggregation of optimal cost from a low-level motion planner, which is achieved by employing a sampling-based motion planner. Furthermore, we propose a sharable sensory data representation i.e. deformed $M_t$ to initialize RL policy model by IL in event-based simulator SUMO. In the experiment, we show that training HBMP policy is very efficient. It outperforms the comparative methods on CARLA benchmark. Finally, the real-world experiment verifies the good generalization ability of HBMP policy.







\bibliographystyle{ieeetr} 
\bibliography{mypaper}

\begin{thebibliography}{10}

\bibitem{paden2016survey}
B.~Paden, M.~{\v{C}}{\'a}p, S.~Z. Yong, D.~Yershov, and E.~Frazzoli, ``A survey
  of motion planning and control techniques for self-driving urban vehicles,''
  {\em IEEE Transactions on intelligent vehicles}, vol.~1, no.~1, pp.~33--55,
  2016.

\bibitem{bojarski2016end}
M.~Bojarski, D.~Del~Testa, D.~Dworakowski, B.~Firner, B.~Flepp, P.~Goyal, L.~D.
  Jackel, M.~Monfort, U.~Muller, J.~Zhang, {\em et~al.}, ``End to end learning
  for self-driving cars,'' {\em arXiv preprint arXiv:1604.07316}, 2016.

\bibitem{codevilla2018end}
F.~Codevilla, M.~Miiller, A.~L{\'o}pez, V.~Koltun, and A.~Dosovitskiy,
  ``End-to-end driving via conditional imitation learning,'' in {\em 2018 IEEE
  International Conference on Robotics and Automation (ICRA)}, pp.~1--9, IEEE,
  2018.

\bibitem{liang2018cirl}
X.~Liang, T.~Wang, L.~Yang, and E.~Xing, ``Cirl: Controllable imitative
  reinforcement learning for vision-based self-driving,'' in {\em Proceedings
  of the European Conference on Computer Vision (ECCV)}, pp.~584--599, 2018.

\bibitem{perot2017end}
E.~Perot, M.~Jaritz, M.~Toromanoff, and R.~De~Charette, ``End-to-end driving in
  a realistic racing game with deep reinforcement learning,'' in {\em
  Proceedings of the IEEE Conference on Computer Vision and Pattern Recognition
  Workshops}, pp.~3--4, 2017.

\bibitem{mukadam2017tactical}
M.~Mukadam, A.~Cosgun, A.~Nakhaei, and K.~Fujimura, ``Tactical decision making
  for lane changing with deep reinforcement learning,'' 2017.

\bibitem{liu2018elements}
J.~Liu, P.~Hou, L.~Mu, Y.~Yu, and C.~Huang, ``Elements of effective deep
  reinforcement learning towards tactical driving decision making,'' {\em arXiv
  preprint arXiv:1802.00332}, 2018.

\bibitem{shalev2016safe}
S.~Shalev-Shwartz, S.~Shammah, and A.~Shashua, ``Safe, multi-agent,
  reinforcement learning for autonomous driving,'' {\em arXiv preprint
  arXiv:1610.03295}, 2016.

\bibitem{dosovitskiy2017carla}
A.~Dosovitskiy, G.~Ros, F.~Codevilla, A.~Lopez, and V.~Koltun, ``Carla: An open
  urban driving simulator,'' {\em arXiv preprint arXiv:1711.03938}, 2017.

\bibitem{behrisch2011sumo}
M.~Behrisch, L.~Bieker, J.~Erdmann, and D.~Krajzewicz, ``Sumo--simulation of
  urban mobility: an overview,'' in {\em Proceedings of SIMUL 2011, The Third
  International Conference on Advances in System Simulation}, ThinkMind, 2011.

\bibitem{schwarting2018planning}
W.~Schwarting, J.~Alonso-Mora, and D.~Rus, ``Planning and decision-making for
  autonomous vehicles,'' {\em Annual Review of Control, Robotics, and
  Autonomous Systems}, 2018.

\bibitem{thrun2006stanley}
S.~Thrun, M.~Montemerlo, H.~Dahlkamp, D.~Stavens, A.~Aron, J.~Diebel, P.~Fong,
  J.~Gale, M.~Halpenny, G.~Hoffmann, {\em et~al.}, ``Stanley: The robot that
  won the darpa grand challenge,'' {\em Journal of field Robotics}, vol.~23,
  no.~9, pp.~661--692, 2006.

\bibitem{montemerlo2008junior}
M.~Montemerlo, J.~Becker, S.~Bhat, H.~Dahlkamp, D.~Dolgov, S.~Ettinger,
  D.~Haehnel, T.~Hilden, G.~Hoffmann, B.~Huhnke, {\em et~al.}, ``Junior: The
  stanford entry in the urban challenge,'' {\em Journal of field Robotics},
  vol.~25, no.~9, pp.~569--597, 2008.

\bibitem{fan2018baidu}
H.~Fan, F.~Zhu, C.~Liu, L.~Zhang, L.~Zhuang, D.~Li, W.~Zhu, J.~Hu, H.~Li, and
  Q.~Kong, ``Baidu apollo em motion planner,'' {\em arXiv preprint
  arXiv:1807.08048}, 2018.

\bibitem{chen2015deepdriving}
C.~Chen, A.~Seff, A.~Kornhauser, and J.~Xiao, ``Deepdriving: Learning
  affordance for direct perception in autonomous driving,'' in {\em Proceedings
  of the IEEE International Conference on Computer Vision}, pp.~2722--2730,
  2015.

\bibitem{pan2017agile}
Y.~Pan, C.-A. Cheng, K.~Saigol, K.~Lee, X.~Yan, E.~Theodorou, and B.~Boots,
  ``Agile autonomous driving using end-to-end deep imitation learning,'' {\em
  arXiv preprint arXiv:1709.07174}, 2017.

\bibitem{sallab2017deep}
A.~E. Sallab, M.~Abdou, E.~Perot, and S.~Yogamani, ``Deep reinforcement
  learning framework for autonomous driving,'' {\em Electronic Imaging},
  vol.~2017, no.~19, pp.~70--76, 2017.

\bibitem{wang2018deep}
S.~Wang, D.~Jia, and X.~Weng, ``Deep reinforcement learning for autonomous
  driving,'' {\em arXiv preprint arXiv:1811.11329}, 2018.

\bibitem{sallab2016end}
A.~E. Sallab, M.~Abdou, E.~Perot, and S.~Yogamani, ``End-to-end deep
  reinforcement learning for lane keeping assist,'' {\em arXiv preprint
  arXiv:1612.04340}, 2016.

\bibitem{pomerleau1989alvinn}
D.~A. Pomerleau, ``Alvinn: An autonomous land vehicle in a neural network,'' in
  {\em Advances in neural information processing systems}, pp.~305--313, 1989.

\bibitem{chen2019learning}
D.~Chen, B.~Zhou, V.~Koltun, and P.~Kr{\"a}henb{\"u}hl, ``Learning by
  cheating,'' {\em arXiv preprint arXiv:1912.12294}, 2019.

\bibitem{kelly2019hg}
M.~Kelly, C.~Sidrane, K.~Driggs-Campbell, and M.~J. Kochenderfer, ``Hg-dagger:
  Interactive imitation learning with human experts,'' in {\em 2019
  International Conference on Robotics and Automation (ICRA)}, pp.~8077--8083,
  IEEE, 2019.

\bibitem{codevilla2019exploring}
F.~Codevilla, E.~Santana, A.~M. L{\'o}pez, and A.~Gaidon, ``Exploring the
  limitations of behavior cloning for autonomous driving,'' in {\em Proceedings
  of the IEEE International Conference on Computer Vision}, pp.~9329--9338,
  2019.

\bibitem{mnih2013playing}
V.~Mnih, K.~Kavukcuoglu, D.~Silver, A.~Graves, I.~Antonoglou, D.~Wierstra, and
  M.~Riedmiller, ``Playing atari with deep reinforcement learning,'' {\em arXiv
  preprint arXiv:1312.5602}, 2013.

\bibitem{silver2017mastering}
D.~Silver, T.~Hubert, J.~Schrittwieser, I.~Antonoglou, M.~Lai, A.~Guez,
  M.~Lanctot, L.~Sifre, D.~Kumaran, T.~Graepel, {\em et~al.}, ``Mastering chess
  and shogi by self-play with a general reinforcement learning algorithm,''
  {\em arXiv preprint arXiv:1712.01815}, 2017.

\bibitem{hosu2016playing}
I.-A. Hosu and T.~Rebedea, ``Playing atari games with deep reinforcement
  learning and human checkpoint replay,'' {\em arXiv preprint
  arXiv:1607.05077}, 2016.

\bibitem{hester2018deep}
T.~Hester, M.~Vecerik, O.~Pietquin, M.~Lanctot, T.~Schaul, B.~Piot, D.~Horgan,
  J.~Quan, A.~Sendonaris, I.~Osband, {\em et~al.}, ``Deep q-learning from
  demonstrations,'' in {\em Thirty-Second AAAI Conference on Artificial
  Intelligence}, 2018.

\bibitem{sauer2018conditional}
A.~Sauer, N.~Savinov, and A.~Geiger, ``Conditional affordance learning for
  driving in urban environments,'' {\em arXiv preprint arXiv:1806.06498}, 2018.

\bibitem{isele2018navigating}
D.~Isele, R.~Rahimi, A.~Cosgun, K.~Subramanian, and K.~Fujimura, ``Navigating
  occluded intersections with autonomous vehicles using deep reinforcement
  learning,'' in {\em 2018 IEEE International Conference on Robotics and
  Automation (ICRA)}, pp.~2034--2039, IEEE, 2018.

\bibitem{fox1997dynamic}
D.~Fox, W.~Burgard, and S.~Thrun, ``The dynamic window approach to collision
  avoidance,'' {\em IEEE Robotics \& Automation Magazine}, vol.~4, no.~1,
  pp.~23--33, 1997.

\bibitem{ross2011reduction}
S.~Ross, G.~Gordon, and D.~Bagnell, ``A reduction of imitation learning and
  structured prediction to no-regret online learning,'' in {\em Proceedings of
  the fourteenth international conference on artificial intelligence and
  statistics}, pp.~627--635, 2011.

\bibitem{schulman2017proximal}
J.~Schulman, F.~Wolski, P.~Dhariwal, A.~Radford, and O.~Klimov, ``Proximal
  policy optimization algorithms,'' {\em arXiv preprint arXiv:1707.06347},
  2017.

\bibitem{abadi2016tensorflow}
M.~Abadi, P.~Barham, J.~Chen, Z.~Chen, A.~Davis, J.~Dean, M.~Devin,
  S.~Ghemawat, G.~Irving, M.~Isard, {\em et~al.}, ``Tensorflow: A system for
  large-scale machine learning,'' in {\em 12th $\{$USENIX$\}$ Symposium on
  Operating Systems Design and Implementation ($\{$OSDI$\}$ 16)}, pp.~265--283,
  2016.

\bibitem{li2018rethinking}
Z.~Li, T.~Motoyoshi, K.~Sasaki, T.~Ogata, and S.~Sugano, ``Rethinking
  self-driving: Multi-task knowledge for better generalization and accident
  explanation ability,'' {\em arXiv preprint arXiv:1809.11100}, 2018.

\end{thebibliography}

\end{document}